\pdfoutput=1
\documentclass{article}

\usepackage{PRIMEarxiv}

\usepackage{multirow}
\usepackage[utf8]{inputenc} 
\usepackage[T1]{fontenc}    
\usepackage{hyperref}       
\usepackage{url}            
\usepackage{booktabs}       
\usepackage{amsfonts}       
\usepackage{nicefrac}       
\usepackage{microtype}      
\usepackage{lipsum}
\usepackage{fancyhdr}       
\usepackage{graphicx}       
\graphicspath{{media/}}     
\usepackage{amsmath} 
\title{PHUDGE: Phi-3 as Scalable Judge}
\author{
  Mahesh Deshwal \\
  \texttt{m.deshwal93@gmail.com} \\
  \AND
  Apoorva Chawla \thanks{Editing \& Proofreading} \\
  \texttt{apoorvachawla@gmail.com} \\
}

\begin{document}
\maketitle

\begin{abstract}
With increase in use of Large Language Models (LLMs), there is an imminent need to automatically evaluate the quality of outputs being produced by these without human intervention. To address this issue, the most preferred approach right now is to use the proprietary LMs such as GPT-4, Claude-3 and Gemini but they have their own set of challenges like 1)Cost of the propriety LMs 2)Exposure of sensitive data to third party 3)Inability to fine-tune on custom rubrics 4)Latency in generation 5)Absence of confidence (probability) for score prediction due to generative nature. Another approach is to build an in-house evaluator LLM for the same but deploying them in live production flow is difficult due to latency issue. In this paper cum technical report, we present PHUDGE \footnote{Code present at: \href{https://github.com/deshwalmahesh/PHUDGE}{https://github.com/deshwalmahesh/PHUDGE}}: A fine-tuned Phi-3 model that achieved SOTA results in 4 tasks (Feedback Test, Feedback OOD, MT Human, Preference Test) surpassing each and every existing model in latency and throughput. It shows very strong correlation not only with GPT-4 but with Human annotators too in unseen data as well as in both absolute and relative grading tasks. \textbf{\textit{We have not only addressed the usage of small LMs for cost effective production grade systems but have also shown that Causal modelling is not only slow in nature but sometimes it can hinder model's learning capabilities and should be replaced by simpler tasks whenever we can to make the overall system faster and better}}. We show that by following systematic ML experimentation, thoughtful data augmentation and re purposing the problem itself, we can even beat 10x bigger models even with lesser training data. To the best of our knowledge, we're the first one to experiment and showcase the usage of generalised version of Earth Mover’s Distance \cite{hou2017squared} (AKA Wasserstein distance) by using Minkowski Distance\cite{10.1007/978-3-642-18129-0_75} with a penalty to control loss smoothing and can be used as a loss function instead of Cross Entropy to get stable training and better results for grading tasks.
\end{abstract}

\section{Introduction}
Evaluating the quality of LLM generated responses has been a tough nut to crack, specially without human intervention of any kind. To address this problem, the most popular approach is to use an LLM as an evaluator in one form or another. On top of it, the data distribution, model capability and biases of LLMs make it look like finding a needle with a needle and can be debated whether it is the best approach to choose. We also agree that it might not be the best (which is Human evaluation) but it is definitely the second best without a doubt - fast, cost effective and relatively accurate judgements.

We have investigated the usage of small (<= 4B) yet powerful LLMs for a fast and cost effective evaluation of responses judging not only propriety models like GPT-4 but also by humans. We extend our experimentation and work to use the absolute quality model in a relative setting to get surprisingly good results for human scoring. More than getting state of the art results on few benchmarks, the core ideology and motive of this paper is to put forth the ideas and effects of using right model, data, right data and correct modelling approach.  Here are some of our contributions from the paper:
\begin{itemize}
\item We present a LoRA \cite{hu2021lora} (rank = 128) fine-tuned on top of Phi-3 model that obtained SOTA results on 4 different benchmarks beating current best models of up to 10x in size,
    \item Through our experiments, we try to address the scoring problem as Causal, Classification and Regression showing how Causal language modelling is not always the best approach and can sometimes be a hindrance to the model's full potential as well.
    \item We address LLM response judgement problem as a traditional classification and show how a modified version of EMD can be used as a replacement of Cross Entropy to get better results. 
    \item  In this paper, we tried to explore some of the debatable questions like Batch Size vs Metric effect, how to choose base model for a task and a simple test to check whether there is a chance of test data leakage.
    \item In the end, we also release the Top-5 contexts (from Wikipedia dumps) for Feedback dataset on 3 different criteria which can be further used for augmentations and RAG on the same data.
\end{itemize}

\section{Related Work}

\begin{figure}
    \centering
    \includegraphics[width=0.85\linewidth]{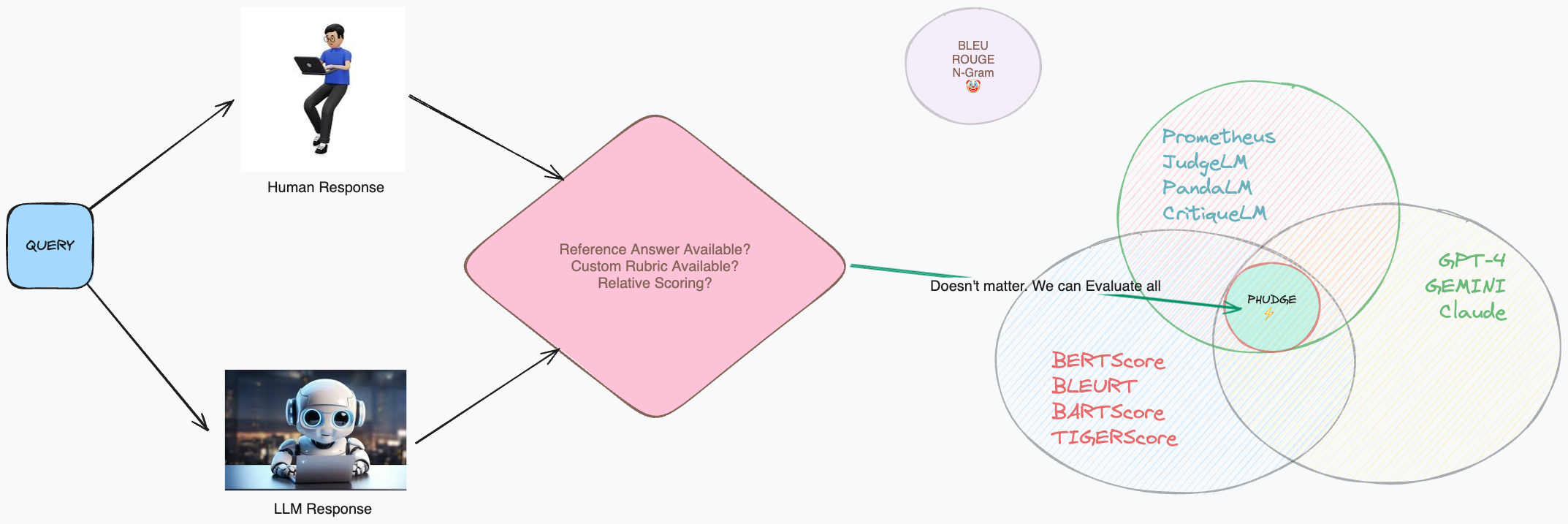}
    \caption{NLG Evaluation Methods}
    \label{fig:PHUDGE_IMG}
\end{figure}

Evaluation of NLG tasks had been going on with the lexical similarity scores like BLEU\cite{papineni-etal-2002-bleu}, ROUGE\cite{lin-2004-rouge} which rely on the N-gram overlapping of the words. With advancement in AI and specially in generative domain, models have become better and better which can skip all the words and still give a correct answer. To capture this semantic meaning on generation some metrics like BERTScore\cite{zhang2020bertscore}, BLEURT\cite{sellam-etal-2020-bleurt}, BARTScore\cite{yuan2021bartscore}, TIGERScore\cite{jiang2023tigerscore} etc. have been proposed which mostly use the embedding semantic similarity of query and response.

One issue with all of the above is the fine grained evaluations on custom logic. Every task has their own set of rules on to which they want to evaluate. With the introduction of idea of using LLM as a Judge\cite{zheng2023judging} which can take custom rules or a reference answer for evaluation, came a promising technique to match the judgement depth \& granularity that human evaluation has. Some great work has been done in this area by numerous researchers through introduction of various evaluation suites and benchmarks such as G-Eval\cite{liu2023geval}, FLASK\cite{ye2024flask}, Chat Eval\cite{chan2023chateval} etc. A broad survey of it can be seen in the report\cite{li2024leveraging}.

As addressed earlier, with the usage of propriety LLMs, there are some definitive issues specially regarding the control, price and latency and to address the security and pricing issue, there have been some promising work done such as PandaLM\cite{wang2023pandalm} which fine tunes a 7B+ models on questions and responses data. One of the major issues with this approach is that it only addresses the Relative scoring part. In real world scenario, we might not have a second answer to compare with. Adding on to this idea, JudgeLM\cite{zhu2023judgelm} came with 3 models of size [7,13,32]B to create custom judges on relative as well as absolute scoring tasks beating PandaLM test data with zero shot. To induce fine grained evaluations which the previous approaches were lacking, Prometheus \footnote{Prometheus(1): \href{https://arxiv.org/abs/2310.08491}{https://arxiv.org/abs/2310.08491}} introduces 100K Feedback Data and fine-tuned models which uses rubric and reference answers which is aligned with real life evaluation  scenario. One major issue with Prometheus was that it only addressed absolute scoring so a new approach Prometheus-2\cite{kim2024prometheus} has been introduced by the authors which extends Feedback Collection Data to create Preference data to add Relative and Absolute scoring to their models via added data and model merging.

\section{Dataset}
 \subsection{Train Data}
For our experiments, we have leveraged the original Feedback Collection dataset first released in the Prometheus paper. We have created a split for 95\% Train and 5\% Test data and all the training in all the experiments is done on this training split only. Reason for making a split is that we did not use the test datasets until unless we were satisfied with the validation data performance and it worked as a guiding criteria for final model selection. Original data is generated by GPT-4 and has 4 inputs as:
\begin{itemize}
    \item Instruction (Original Question asked by user)
    \item  Response to Evaluate (Response given by a LLM)
    \item Customised Score Rubric (Diverse set of rules on which the scoring should be given)
    \item Reference answer (What should be the ideal answer to the question)
\end{itemize}
On the output side, it has 2 components which are the responses from GPT-4 when used as an evaluator:
\begin{itemize}
    \item Feedback (The reason given for the high/low score which should refer to the rubric to justify the feedback) 
    \item Score: An integer score (one of 1,2,3,4,5. All the Scores are occurring with almost same frequency so there is no label/score bias in data )
\end{itemize}
\subsection{ Benchmark Test Data \& Metrics}
For the evaluation data, there 9 different datasets used, details of which are given below.  Metrics used for the absolute scoring are different correlation mainly as Pearson, Kendall and Spearman. Along with that, we have also tested our models on R-Squared, MAE and MSE. For relative scoring, metric used is Accuracy.

\begin{tabular}{ |p{4.5cm}|p{1.5cm}|p{2.7cm}|p{2cm}|p{2cm}|  }
 \hline
 \multicolumn{4}{|c|}{Eval Dataset} \\
 \hline
 Name & Samples & Absolute / Relative & Scorer: GPT / Human & Ref Ans: Y/N\\
 \hline
 Feedback Collection Test   & 1000    &A&   G & Y\\
 Feedback Collection OOD &  1000  &  A & G & Y\\
 Vicuna & 300 & A&  G & Y\\
 Flask  & 2000 & A&  BOTH & Y\\
 Preference Collection Test OOD &  2000  & R & G & Y\\
 HHH& 200  & R   &H & N\\
 MT Human& 1400  & R& H & N \\
 PandaLM & 2500 & R & H&  N \\
 JudgeLM & 5000 & R & G & N \\
 \hline
\end{tabular}

\section{Methodology and Experiments}
Though we conducted experiments on couple of small LLMs including Phi-2 (2.8B)\cite{li2023textbooks}, Gemma (2B)\cite{gemmateam2024gemma} and Qwen (4B)\cite{bai2023qwen} for initial experiments, we found Phi-3 (3.8B) a much better model for our task and will be talking about it for the rest of the report. For our Training setup, we have kept most of the hyper-parameters same for all the models except Batch Size where we wanted to see the effect of metrics on batch size. Apart from using the Pearson, Kendall and Spearman correlation for Absolute scoring, we used R Squared, Mean Absolute Error and Mean Squared Error though other papers have not used this criteria in evaluation but these can be great metric to quantify prediction errors. On the Relative scoring side, we only used Accuracy score as in other papers.

\subsection{Base Model Selection \& Test Data Leakage Test}
To select the base model, apart from roughly looking at the multiple public benchmarks data, one thing we looked at was the performance of our 5\% validation split on the base model. We did this exercise for two main reason being the arguable leakage of benchmark data in base model and to get a sense of how our base model performs on the validation data (which is training data in original Prometheus paper). Being a sceptic, we also had the question whether there is a leakage of test data with Phi-3 because it shows very good results \ref{section:eval} on some of the datasets. One possible explanation to this is that Phi-3 came \textit{before} the release of Prometheus-2 paper and as we were already working on original Feedback Data so we got another benchmark to test the model's generalisation capabilities with zero shot. To double down on the idea, we tested base model results with fully fine-tuned Causal one and similarly final Regression model with the one where we froze all model weights and fine-tuned only the regression head. Comparing it with increase in LoRA rank gives a hint that the evaluation property is learned by the model through data and fine tuning rather than test data overfitting.

\subsection{Re-thinking Problem Definition}
Causal Language Modelling is the go to approach for LLMs and specially for this task as the original data has a feedback component which gives a reason of why the score was given high or low. We tested our initial models with a small rank LoRA (8) to see the results on validation set. However, looking at this problem differently, you'll notice that there are some issue with using Causal as:
\begin{itemize}
    \item Training and specially inference is slower because of the auto regressive nature
    \item In the generative nature, we might not always get the final result we want as the model might not predict [RESULT] [SCORE]  within the token generation limit (which is 256 in our case) or maybe not at all.
    \item One more issues is the confidence on the prediction. Even looking at the token probability for score for the final token is not a very accurate idea of model's confidence in prediction.
    \item We lose the fine grained nature of the scores. The lowest non zero error rate we will get between 2 prediction will be 1 as the model will be trained to predict the integers as 1,2,3,4,5
\end{itemize}

Looking at these issues, we need to re think whether using generation is really necessary. Although, in tasks where we care more about the score than the reason for scoring itself, Regression or Classification is definitely a better approach. Also, we only trained 2 models for Causal which is LoRA-128 and full fine tuning and both of those do not use our augmentations which we applied for Regression models only. Full Fine-tuned version got a Pearson Correlation of 0.91 just matching the current best model available.

\subsubsection{Classification}
\begin{figure}
    \centering
    \includegraphics[width=0.75\linewidth]{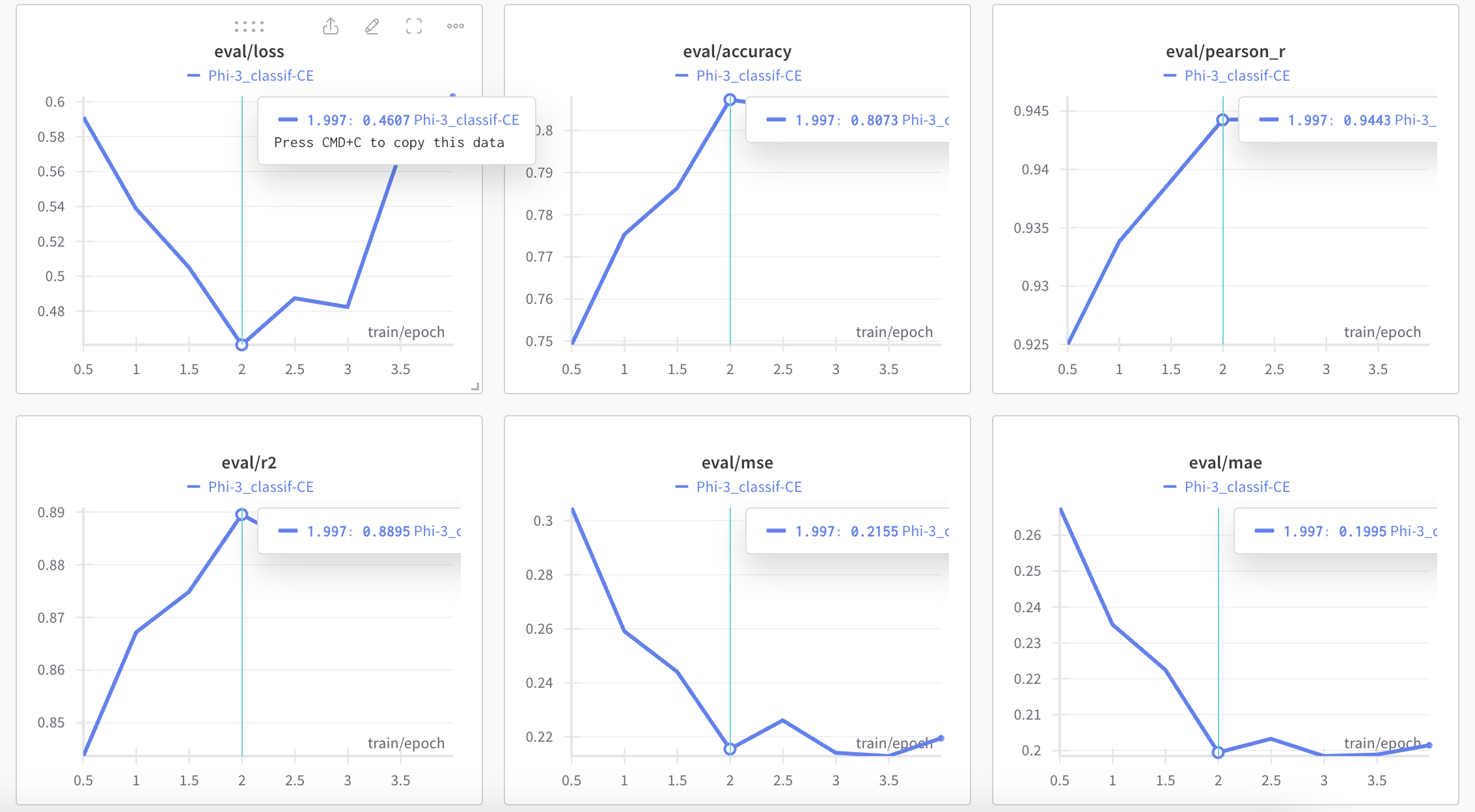}
    \caption{Training With CE loss (Early Overfitting)}
    \label{fig:ce_training}
\end{figure}

Treating the problem as a classification gives us better control on the confidence and we can tweak the acceptance threshold but this comes with a cost of losing the feedback of why the model gave a certain score. Iterating further on this, we used a Multi Class (5) classification using the Cross Entropy loss which gave us a highest ever results for all the experiments tried till that time but we that noticed that CE model starts overfitting very early \ref{fig:ce_training}  and even with all the correlations being higher, the R-Squared and MSE were not up to the mark and we sensed that the issue might be in the working of CE Loss. It treats all the misclassifications as equal but in our case, predicting 2 as 3 is relatively better than predicting 5. This problem is better handled by the modified EMD loss as it uses Cumulative Density Function (CDF). So by leveraging the difference between cumulative sum of two probability distributions (GT and Pred), we use it as a loss function for next experiments.

We have implemented 3 versions of EMD, Squared using a simple Squared EMD as given in equation \ref{eq_emd_squared} , Squared + Summed which sums the class wise error for the batch before taking the batch mean and the third, which is a generalised and flexible approach as described in \ref{gen_minow_eq}. Lets us say \(p\) and \(t\) be two distributions (our predicted and gt probabilities) which in ordered classification tasks, satisfy all the 3 conditions of sorted, equal mass and 1-D. So in the closed form, EMD can be computed using CDF as:

\begin{equation}
\operatorname{EMD}(\mathbf{p}, \mathbf{t})=\left(\frac{1}{C}\right)^{\frac{1}{l}}\|\operatorname{CDF}(\mathbf{p})-\operatorname{CDF}(\mathbf{t})\|_l
\end{equation}

Setting \(\textit{l }\)as 2 ( Euclidean distance) and dropping the norm term in calculating distance, the equation is converted to modified Squared EMD Loss as:

\begin{equation}
\label{eq_emd_squared}
\mathrm{E}_L(\mathbf{p}, \mathbf{t})=\sum_{i=1}^C\left(\operatorname{CDF}_i(\mathbf{p})-\mathrm{CDF}_i(\mathbf{t})\right)^2
\end{equation}

where \(\operatorname{CDF}_i(\mathbf{p})\) is the i-th element of  \(\mathbf{p}\)

We experimented by introducing a parameter \(\operatorname\alpha\) in the flexible version which can be used to control the smoothing of error and can be thought of as an extension of Minkowski distance and generalised norms. 

\begin{equation}
\label{gen_minow_eq}
EMD_s(x, y)=\left(\sum_{i=1}^n\left|x_i-y_i\right|^p\right)^{\frac{\alpha}{p}}
\end{equation}

where \(EMD_s(x, y)\) is the EMD loss for the \(\operatorname{s}\)-th sample in the batch, \(\operatorname{n}\) = 5 classes (scores = 1,2,3,4,5) and \(\operatorname{x}\), \(\operatorname{y}\) are the Cumulative Sum (1-D Vectors) of Softmaxed logits and One-Hot-Encoded GT labels of that sample respectively in our use case.

If \(\operatorname\alpha\) is equal to \(\operatorname{l}\) = 2, then it will act as the Squared EMD (L-2 Norm). Individual sample losses are in range [0,1] so tweaking the value \( 0 < \operatorname\alpha\) < \(\operatorname{l}\)  or  \( \operatorname\alpha\) > \(\operatorname{l}\) will increase or relax the overall loss accordingly as needed for a particular task.

\begin{figure}
    \centering
    \includegraphics[width=0.75\linewidth]{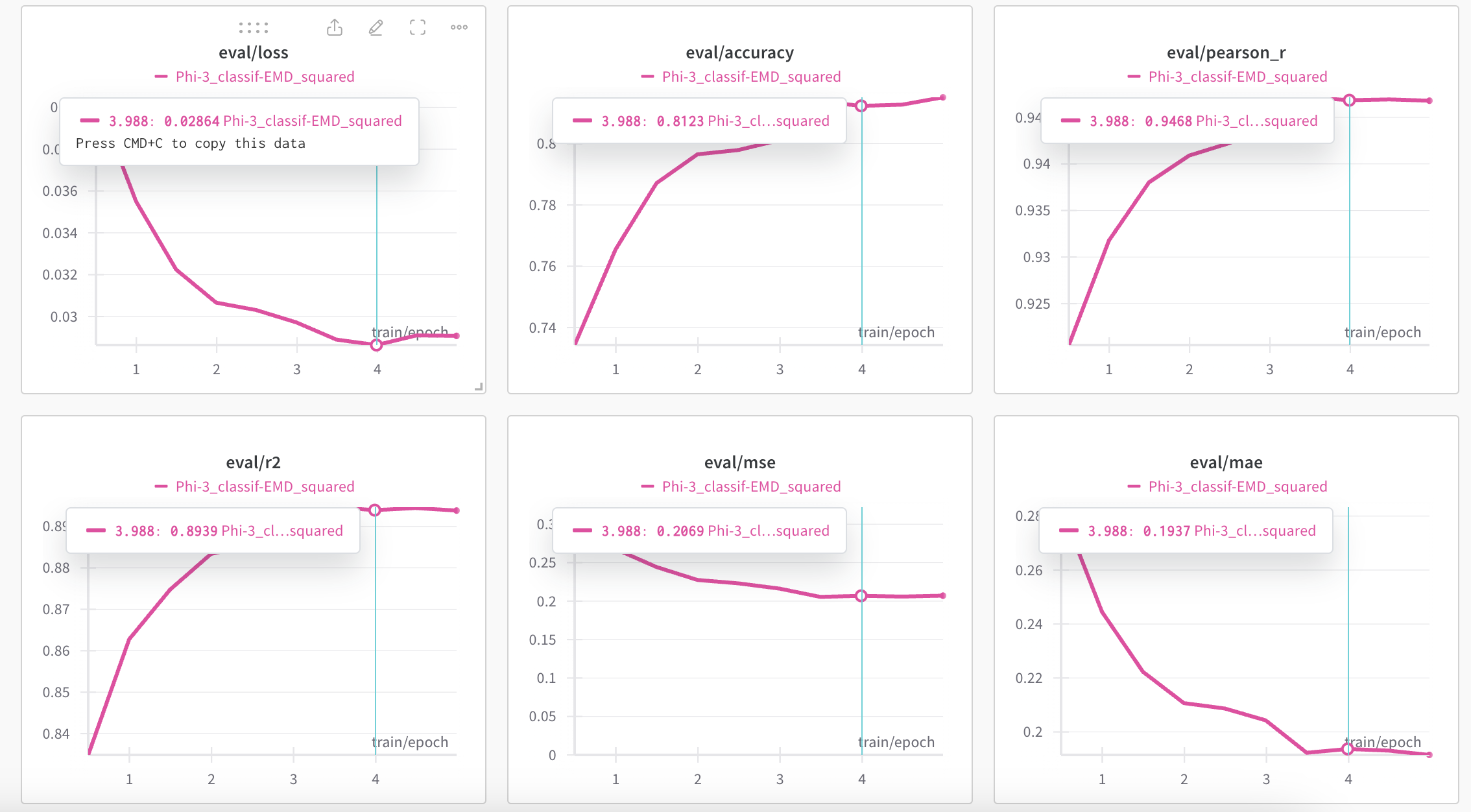}
    \caption{Training with EMD Loss}
    \label{fig:emd_training}
\end{figure}

Experimenting with EMD and Squared EMD, we found that squared version is a bit slower than CE but is more stable and yields better results \ref{fig:emd_training} specially in terms of MSE. EMD loss did not show clear signs of overfitting till the last epoch while for the CE, validation loss started increasing almost halfway training \ref{fig:ce_training}.

\subsubsection{Regression}
Though we can have probabilities for the scores but classification still lacks floating score values and this can contribute to a different MAE, MSE values than expected. The reason of our classes being ordered classes (scores), we used EMD instead of CE and the same thing led us to test with regression as a problem. Regression is usually avoided due to it's unstable and hard to optimise nature specially in the case of biased labels. But because our labels/scores are uniformly distributed and are within a very narrow range of 1-5, thanks to the data creation process, testing the problem with MSE as a loss function led to highest correlation and least MAE, MSE. 

\begin{figure}
    \centering
    \includegraphics[width=0.9\linewidth]{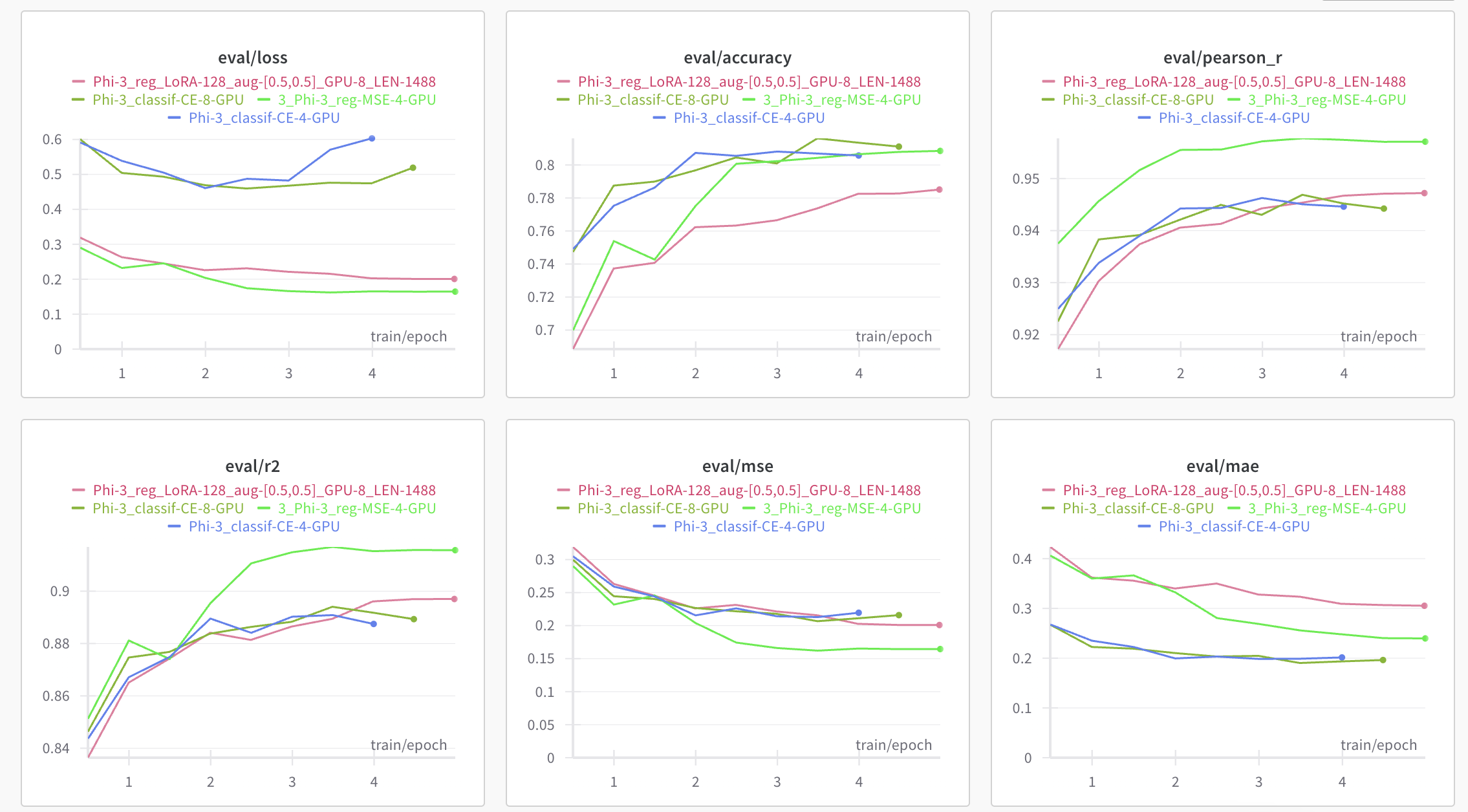}
    \caption{Effects of Batch and modelling approach}
    \label{training_graph}
\end{figure}

There is also one more reason to use Regression in place of classification as the classification models started overfitting after 2 epochs while the Regression models kept on going without an issue and there is still a room for improvement for further couple of epochs as can be seen in Figure \ref{training_graph}. For all of our future iterations, specially augmentation, we kept regression as a choice. All the scores that we present in the paper are from the Regression model unless explicitly mentioned.

\subsection{Ablation Study}
In all the experiments we have used DeepSpeed ZeRO-2 with Multi GPU as our model could fit in single GPU with gradient checkpointing enabled. We did not use qLoRA in any of the experiments because we did not need to and fully fine-tuned only one Causal Model (which had a Feedback data correlation of 0.91). Our code can be easily used on a single GPU too. Apart from few, we mostly kept all the hyper-parameters constant throughout the process. Refer to Table~\ref{tab:hyper_param_table} for the hyper-parameter values that we have used in our experiments.

\begin{table}
 \caption{Hyper-parameters}
  \centering
  \begin{tabular}{lll}
    \cmidrule(r){1-2}
    Name     & Value  \\
    \midrule
    Sequence Length &  1440, 1652\\
    Batch Size & Variable size that utilized the max GPU given other params \\
    LoRA Rank&  8, 128, Full-Finetuning\\
    qLoRA &  No\\
    Epochs&  1,3,5\\
    Gradient Check-pointing &  Yes\\
    Gradient Accumulation steps &  1\\
    LR    & 2e-5\\
    Scheduler & Cosine\\
    Warmup Ratio & 0.05\\
   Optimizer &  AdamW\\
   Weight Decay &  0.005\\
    GPU     & A10 \\
    No. of GPU     & 1,4,8\\
    BF-16     & Yes\\
    Torch Compile     & Yes\\
    Attention Type     & Flash-Attention-2\\
    \bottomrule
  \end{tabular}
  \label{tab:hyper_param_table}
\end{table}

\subsubsection{Effect of Batch Size, LoRA and Sequence Length \ref{training_graph}}
Apart from LoRA rank, which had the highest effect on the scores, Batch size also had a noticeable effect followed by Sequence Length on the validation scores. We used a batch size which fully utilized the GPU usage given LoRA Rank, No of GPU and Sequence Length.  We noticed that increasing batch size to almost double while keeping everything same resulted in worse performance. Similarly, doing the same for Regression resulted in a degraded performance. Note that we kept the a maximum of 5 epochs (except for full fine tuning which is 3) and declared the model best where the validation loss was minimum. Increasing the batch size also controlled the increase in validation loss during training.

Context length of Phi-3 is quite high but we kept the Sequence Length at the 99 percentile which is different for Classification and Regression (1488) versus Causal (1656) where Feedback is part of of the input. We noticed that using a higher sequence length (ex 1656) at prediction time even when the model is trained on 1488 increases the performance Feedback OOD by almost 1\% in Regression tasks.  

Rank of LoRA is the most important hyper-parameter for the task. Training only the Regression head without LoRA (but frozen model weights) was the worst while increasing the rank led to increase in performance. Even though we did not fully fine tune any model for Regression or Classification, we feel that doing that would not be beneficial for these tasks as it might lead to overfitting pretty easily after a certain rank. We leave the higher rank experimentation for future work.

One experiment we did specially with the Causal model to compute the loss on the Whole Generation versus Feedback only and noticed that the Whole Generation model performed poorly compared to the generation only scenario where we masked the input instructions in loss computation and calculated loss only for the Feedback generation. It can be done easily by providing a token id \textit{[32001]} to the tokenizer which signifies the start of generation from  \textit{<|assistant|>}.

\subsection{Augmentations}
Most generalised model is the one which is trained on the augmented version of the inputs. We used the original data and dropped Reference Answer and Score Rubric randomly with 50\% probability. Note that these two augmentations are independent of each other so for some training sample, both the rubric and reference answer can be missing at once. Out of all the regression models, the one trained on the augmentation turned out to be the most versatile when tested with reference free, rubric free and reference AND rubric free settings which we have used for our final comparisons. 

One thing to be noted is that while this model is performing better overall for both seen and OOD data, it's performance is slightly lesser (~1\%) in Feedback Test which is a seen rubric data. This comparison is done with a model which was trained with no augmentation on the same data and after training. It supports our statement that without augmentation the model becomes too specialised at current data only and the robust judgement attribute is acquired mostly due to the augmentation. We have also created and released reference Wikipedia data per row having 5 top contexts. For each row, there are 3 different contexts depending on the context used for similarity with paragraphs which are Instruction Only, Response Only and Instruction + Response together. We leveraged \href{https://huggingface.co/thenlper/gte-large}{GTE Large} for embedding creation. Using this data during fine tuning, we can further make the model robust for RAG approach and leave it for future work.

\section{Evaluation and Results}
\label{section:eval}
Out of 4 datasets used, fine-tuned Phi-3 has outperformed in 2 of those with good margin showing very high correlation with GPT-4. We are comparing our results only with the current best model on these tasks. All the results displayed here are Pearson Correlation. All the models except ours are full fine-tuned versions on their own respective datasets. Refer to Table~\ref{tab:absolute_score} for Absolute scoring and Table~\ref{tab:relative_score}, \ref{tab:relative_score_panda_judge} for relative scoring.

\begin{table}
    \centering
    \begin{tabular}{|c|c|c|c|c|c|} \hline  
         Model&  \multicolumn{5}{|c|}{Data}\\ \hline  
         &  Feedback (Test)&  Feedback (OOD)&  Vicuna&  Flask (GPT)& Flask (Human)\\ \hline  
         Prometheus-2 (7x8B)&  -&  0.9&  \textbf{0.68}&  \textbf{0.66}& \textbf{0.55}\\ \hline  
         Prometheus-2 (7B)&  -&  0.88&  0.64&  0.64& 0.54\\ \hline  
         Prometheus (13B)&  0.86&  0.86&  0.49&  0.46& 0.44\\ \hline  
         Prometheus (7B)&  0.86&  0.85&  0.48&  0.35& 0.35\\ \hline  
         Ours&  \textbf{0.93}&  \textbf{0.95}&  0.41&  0.38& 0.37\\\hline 
    \end{tabular}
    \caption{Absolute scoring Benchmark results}
    \label{tab:absolute_score}
\end{table}

\begin{table}
    \centering
    \begin{tabular}{|c|c|c|c|} \hline  
         Model&  \multicolumn{3}{|c|}{Data}\\ \hline  
         &  HHH&  MT Bench Human&  Preference Bench\\ \hline  
         Prometheus-2 (7x8B)&  \textbf{0.85}&  0.55&  0.91\\ \hline  
         Prometheus-2 (7B)&  0.79&  0.56&  0.92\\ \hline  
         Prometheus (13B)&  0.79&  0.58&  -\\ \hline  
         Prometheus (7B)&  0.8&  0.55&  -\\\hline \hline  
         Ours&  0.83&  \textbf{0.59}&  \textbf{0.97}\\\hline 
    \end{tabular}
    \caption{Relative scoring Benchmark results (Accuracy)}
    \label{tab:relative_score}
\end{table}

\begin{table}
    \centering
    \begin{tabular}{|c|c|c|c|} \hline  
         Model&  \multicolumn{3}{c}{Data}\\ \hline  
         &  JudegeLM&  JudgeLM (with Ref)&  PandaLM\\ \hline  
         Prometheus (13B)&  \textbf{0.84}&  \textbf{0.85}&  \textbf{0.68}\\ \hline  
         Prometheus (7B)&  0.81&  0.84&  0.65\\
 PandaLM (7B)& 0.68& 0.63&0.59\\
 GPT-3.5& 0.73& 0.71&0.62\\\hline\hline \hline  
         Ours&  0.72&  0.77&  \textbf{0.68}\\\hline 
    \end{tabular}
    \caption{Relative scoring on JudgeLM and PandaLM
    (Accuracy)}
    \label{tab:relative_score_panda_judge}
\end{table}

For the Relative scoring, the model got best results in 2 datasets out of 5 used. We only trained on 95\% of Feedback Collection and tested it on JudgeLM and PandaLM test sets which have very different distributions of questions and answers in the pairwise setting. Phi-3 showed  pretty high accuracy on JudgeLM Test data with 77\% Accuracy on without reference answer and 80\% on with reference. The small gap in with and without on unseen data supports the generalised nature of our approach. 

\section{Limitations}
There are some known limitations to this approach, major being that we lose the reasoning/feedback of why the score was given. Researchers usually make the model generate feedback \textit{before} the score so as to generate the score based on reasoning and not the other way around. So this approach cannot be used where we care about feedback more than the score for example human assisted quality checking. On the cost of making model faster, we had to let go of this feature even though we have shown that it sometimes acts as a hindrance to the model's true capability (Fully fine-tuned causal model achieving 91\% Accuracy on Feedback test versus LoRA(128) Regression achieving 95\% on the same).

In terms of training time, Squared EMD is a bit slower than CE and the generalised one turned out to be the slowest among all, taking almost 3 seconds more per iteration than CE loss. Even though we have introduced the penalty term \(\operatorname{\alpha}\) in the generalised EMD calculation, still a thorough experimentation is needed to be done in order to truly understand the effects of this in terms of stability and performance on a wide array of scoring tasks.

\section{Conclusion \& Future Work}
The reason for high score on the datasets does not seem to be coming from leakage. We suspect that Phi-3 might have been trained using GPT-4 generated data thus using the preference bias as a positive thing as it's well versed with the data distribution. We can use this idea as a base for future works to find which foundation models have been trained using which proprietary models (like Claude, GPT-4, Gemini etc.) when data details are not released. We have empirically found that using any kind of reference (rubric, reference answer, related context) is almost always beneficial and the more context the judge model has, the better it'll be able to score. So extending the same idea, we have added the Wikipedia contexts to the original data specially for the purpose of making it robust during RAG deployment which is a work for future experiments.

\newpage
\bibliographystyle{unsrt}  
\bibliography{references}

\end{document}